\documentclass[10 pt, journal]{IEEEtran}

\usepackage{graphicx}
\usepackage{verbatim}
\usepackage{amsmath}
\usepackage{eepic}

\begin{document}

\title{Entropy Message Passing}

\author{Velimir M. Ili\'c,
        Miomir S. Stankovi\'c
        and Branimir T. Todorovi\'c
\thanks{V. Ili\'c is with the Department of Informatics, Faculty of Sciences and Mathematics, University of Ni\v s,
Serbia,}
\thanks{M. Stankovi\'c is with the Faculty of Occupational Safety, University of Ni\v s,
Serbia,}
\thanks{B. Todorovi\'c is with the Department of Informatics, Faculty of Sciences and Mathematics, University of Ni\v s,
Serbia.}}

\maketitle


\begin{abstract}
The paper proposes a new message passing algorithm for cycle-free
factor graphs. The proposed "entropy message passing" (\emph{EMP})
algorithm may be viewed as sum-product message passing over the
entropy semiring, which has previously appeared in automata
theory. The primary use of \emph{EMP} is to compute the entropy of
a model. However, \emph{EMP} can also be used to compute
expressions that appear in expectation maximization and in
gradient descent algorithms.
\end{abstract}



\section{Introduction}

The efficient marginalization of a multivariate function is
important in many areas including signal processing, artificial
intelligence, and digital communications. When a cycle-free factor
graph representation of the function is available, then exact
marginals can be computed by sum-product message passing in the
factor graph \cite{Kschischang: FG and SPA}-\cite{Aji-McEliece:
GDL}. In fact, a number of well-known algorithms are special cases
of sum-product message passing.

The "sum" and the "product" in sum-product message passing may
belong to an arbitrary commutative semiring \cite{Aji-McEliece:
GDL},\cite{Wiberg: disertation}. In this paper, we propose to use
it with the entropy semiring and the resulting algorithm will be
called "entropy message passing" (\emph{EMP}). The entropy
semiring was introduced by Cortes et al. \cite{Cortes: KLD betwen
automata} to compute the relative entropy between probabilistic
automata. In this paper, we translate the ideas of \cite{Cortes:
KLD betwen automata} into the language of factor graphs and
message passing algorithms.

The primary use of \emph{EMP} is to compute the entropy of a model
with a cycle-free factor graph for fixed observations. The main
prior work on this subject is by Hernando et al. \cite{Hernando:
Entropy of HMM}; again, a main point of the present paper is to
clarify and to generalize this prior work by reformulating it in
terms of sum-product message passing. However, \emph{EMP} can also
be used to compute expressions that appear in expectation
maximization and in gradient ascent algorithms \cite{Eckford: MUD
with GM}-\cite{Loeliger: Some remarks on FG}; this connection
appears to be new.

The paper is structured as follows. In Section II, we review
sum-product message passing over a commutative semiring. In
Section III, we introduce the entropy semiring. The \emph{EMP}
algorithm is described in Section IV and the mentioned
applications are described in Section V.


\section{Factor Graphs and the Sum-Product Algorithm}

Let \textit{f} be a real multivariate function that depends on the
set of variables $\mathbf{x}=\{x\}_{n=1}^N $ and satisfies
\begin{equation}
\label{Factor rule for f}
f(\mathbf{x})=\prod_{m\in{\mathcal{M}}}f_m(\mathbf{x}_m)
\end{equation}
for a set of indexes $\mathcal{M}$. In the expression (\ref{Factor
rule for f}) each factor $f_m(\mathbf{x}_m)$ depends on
$\mathbf{x}_m \subset \mathbf{x}$ and the subsets $\mathbf{x}_m$
cover $\mathbf{x}$. The factorization (\ref{Factor rule for f})
can be graphically represented by a factor graph
\cite{Kschischang: FG and SPA}-\cite{Bishop: book}. %
A factor graph consists of the variable nodes (drawn as circles),
the factor nodes (drawn as squares) and the connections between
the nodes, where the variable node \textit{n} and the factor node
\textit{m} are connected if and only if the factor $f_m$ depends
on the variable $x_n$. %
An example of a factor graph is given in
Fig. 1.
\begin{figure}[b]
\center
\setlength{\unitlength}{1mm}
\begin{picture}(70, 45)
  \put(3, 8){\circle{6}}
  \put(2, 0){$4$}

  \put(3,11){\line(0,10){13}}

  \put(2, 24){\rule{2mm}{2mm}}
  \put(1, 31){$D$}

  \put(4,26){\line(4,3){12.1}}

 \put(19, 36){\circle{6}}
 \put(18, 41){$1$}

  \put(19,33){\line(0,-1){13}}

  \put(18,18){\rule{2mm}{2mm}}
  \put(17,13){$A$}

  \put(34, 24){\rule{2mm}{2mm}}
  \put(34, 31){$C$}

  \put(34,26){\line(-4,3){12.1}}

  \put(35, 8){\circle{6}}
  \put(34, 0){$3$}

  \put(35,11){\line(0,10){13}}

  \put(36,26){\line(4,3){12.1}}

 \put(51, 36){\circle{6}}
 \put(50, 41){$2$}

  \put(51,33){\line(0,-1){13}}

  \put(50,18){\rule{2mm}{2mm}}
  \put(49,13){$B$}

  \put(66, 24){\rule{2mm}{2mm}}
  \put(66, 31){$E$}

  \put(66,26){\line(-4,3){12.1}}

  \put(67, 8){\circle{6}}
  \put(66, 0){$5$}

  \put(67,11){\line(0,10){13}}
\end{picture}
\center Fig. 1.\ \  The factor graph that corresponds to the
factorization $f_A(x_1)f_B(x_2)f_C(x_1,x_2,x_3)
f_D(x_1,x_4)f_E(x_2,x_5)$.
\end{figure}

The following two problems are of interest in many applications
\cite{MacKay: book}:

\subsubsection {The marginalization problem}
\begin{equation}
\label{Zn problem} Z_n(x_n)=\sum_{\mathbf{x}\setminus{x_n}}
\prod_{m \in \mathcal{M}}f_m(\mathbf{x}_m),
\end{equation}
\noindent where $\sum_{\mathbf{x}\setminus{x_n}}$ denotes the
summing all variables from $\mathbf x$ except $x_n$ and

\subsubsection {The normalization problem}
\begin{equation}
\label{Z problem} Z=\sum_{\mathbf{x}} \prod_{m \in
\mathcal{M}}f_m(\mathbf{x}_m).
\end{equation}

The solution of the second problem can easily be obtained from the
solution of the first problem by means of:
\begin{eqnarray}
\label{Solution for Z problem} Z=\sum_{x_n} Z_n(x_n);
\end{eqnarray}
therefore, in the following paragraphs, we are concerned with the
solution of the first problem.


The marginalization problem (\ref{Zn problem}) can efficiently
been solved using the sum-product algorithm (\emph{SPA})
\cite{Kschischang: FG and SPA}-\cite{Bishop: book}. The
sum-product algorithm operates by passing messages along the edges
of the factor graph of a function to be marginalized. The computed
marginal value is exact for a cycle-free factor graph, but the
algorithm can also be applied on graphs with cycles, in which case
an approximate solution may be obtained \cite{Yedidia: Free energy
and GBP}-\cite{Weiss: Correctnes of LBP}. In this paper, we
consider only cycle-free (i.e., tree structured) factor graphs.

There are two types of messages:

\begin{enumerate}

\item the messages $q_{n\rightarrow m}(x_n)$ from variable to factor nodes and
\item the messages $r_{m\rightarrow n}(x_n)$ from factor to variable nodes,

\end{enumerate}
where the variable and factor nodes participating in the message
passing process are denoted with $n$ and $m$. Note that both types
of messages are functions of the variable that is represented by
the involved node.

The messages are initialized to $q_{n\rightarrow m}(x_n) =1$ and
$r_{m\rightarrow n}(x_n)=f_m(x_n)$, for all variable nodes $n$ and
factor nodes $m$ in the leaves of the factor graph, for all
possible values $x_n$. After that the messages are passed toward
the root that corresponds to the variable for which the marginal
value is computed. The message from a node to its parent is
computed after the messages from all descendants are received,
according to the following rules:

\begin{equation}
\label{Variable -> factor message}
 q_{n\rightarrow m}(x_n)=
 \prod_{m'\in \mathcal{N}(n)\setminus m}r_{m'\rightarrow n}(x_n)
\end{equation}
and
\begin{equation}
\label{Factor -> variable message}
 r_{m\rightarrow n}(x_n)=
 \sum_{\mathbf{x}_m\setminus x_n} f_m(\mathbf{x}_m) \prod_ {n'\in \mathcal{N}(m)\setminus
 n}q_{n'\rightarrow
 m}(x_{n'}).
\end{equation}

Here, $\mathcal{N}(n)\setminus m$ denotes all the nodes that are
neighbors of the node $n$ except for the node $m$, and
$\sum_{\mathbf{x}_m\setminus x_n}$ denotes a sum over all the
variables $\mathbf{x}_m$ that are arguments of $f_m$ except $x_n$.
The process is terminated at the root, where the marginal function
is computed according to:
\begin{equation}
\label{Termiantion} Z_n(x_n)= \prod_ {m \in \mathcal{N}(n)}
r_{m\rightarrow n}(x_n).
\end{equation}
\par

So far, we have implicitly assumed that the function to be
marginalized has as codomain the set of real numbers obtained with
the standard operations $+$ and $\times$. Nevertheless, the
algorithm still works when the codomain is an commutative semiring
(see the next section for the definition). The generalized form of
the algorithm can be obtained straightforwardly by replacing the
operations $+$ and $\times$ from the set of real numbers with the
operations $\oplus$ and $\otimes$ from the semiring
\cite{Kschischang: FG and SPA},\cite{Aji-McEliece:
GDL},\cite{Wiberg: disertation}.


\section {The Entropy Semiring}

In this section we introduce the algebraic notions that will be
useful for development of the \emph{EMP} algorithm.

\emph{Definition 1} \cite{Kuich: book- Semirings} : The system
$\left\langle \ \mathcal{K},\ \oplus,\ \otimes,\ \overline 0, \
\overline 1 \ \right\rangle \ $ is called a commutative semiring
if:

\begin{enumerate}

\item The operations $\oplus$ and $\otimes$ are associative and
commutative;

\item The equalities $k \oplus \overline 0=k$ and $k \otimes \overline 1=k$
hold for all $k \in \mathcal{K}$;

\item The operation $\otimes$ distributes over $\oplus$, i.e., for all
$a,b,c \in \mathcal{K}$ the following equalities hold:

\begin{equation}(a \oplus b) \otimes c = (a \otimes c) \oplus ( b \otimes c),\end{equation}
\begin{equation}c \otimes (a \oplus b) = (c \otimes a) \oplus ( c \otimes b).\end{equation}

\end{enumerate}


Some common commutative semirings are the sum-product semiring
$\left\langle\mathcal{R_+}, + ,\times , 0,  1\right\rangle$, the
Boolean semiring $\left\langle\{0, \ 1\},\wedge,\vee, 0,
1\right\rangle$ and the max-product semiring
$\left\langle\mathcal{R_+},max ,\times, 0, 1 \right\rangle$, where
$\mathcal{R}$ denotes the set of real numbers. Other semirings
used in the message passing algorithms can be found in
\cite{Aji-McEliece: GDL}. In this paper we consider the entropy
semiring \cite{Cortes: KLD betwen automata} (also called the
expectation semiring \cite{Eisner: Parameter Estimation for
PFST}).

\emph{Definition 2} The entropy semiring is a tuple $$\left\langle
\ \mathcal{R}^2,\ \oplus,\ \otimes, \ ( 0 ,\ 0 ), (  1 ,\ 0)\
\right\rangle ,$$ where the operations $\oplus$ and $\otimes$ are
defined with:
\begin{equation}
\label{Entropy semiring- addition}
(x_1,y_1)\oplus(x_2,y_2)=(x_1+x_2,y_1+y_2);
\end{equation}
\begin{equation}
\label{Entropy semiring- multiplication}
(x_1,y_1)\otimes(x_2,y_2)=(x_1x_2,x_1y_2+x_2y_1),
\end{equation}
for all $(x_1,\ y_1),(x_2,\ y_2)\in \mathcal{R}^2$.

The following lemma will be useful for the derivation of the
\emph{EMP} algorithm in the next section.

\newtheorem{lemma}{Lemma}

\begin{lemma}
Let $\mathcal{M}$ be a finite set of indices. Then, for all
$(a_m,\ b_m)\in \mathcal{R}^2, \ m \in \mathcal{M}$ the following
equality holds:

\begin{equation}
\label{Lemma 1.} \bigotimes_{m\in {\mathcal{M}}}(a_m,\ b_m)=
\left(\prod_{m\in \mathcal{M}}a_m,\sum_{m\in
\mathcal{M}}b_m\prod_{j\in \mathcal{M}\setminus \{m\}}a_j\right).
\end{equation}

\begin{IEEEproof}
We prove the lemma by induction over the cardinality of
$\mathcal{M}$. Without loss of generality, suppose that the sets
$\mathcal{M}$ have the form $\{1,\ 2,\ ... \ ,\ k\}$ where $k$ is
from the set of the natural numbers. If $\mathcal{M}$ has two
elements, the equality (\ref{Lemma 1.}) reduces to the definition
for multiplying in an entropy semiring:
$$(a_1,\ b_1)\otimes(a_2,\ b_2)=(a_1a_2,\ a_1b_2+a_2b_1).$$

Now, let the equality hold for some $k$ - element set
$\mathcal{M}_k=\{1,\ 2,\ ... \ ,\ k\}$:
$$\bigotimes_{m\in {\mathcal{M}_{k}}}(a_m,\ b_m)=\left(\prod_{m\in \mathcal{M}_k}a_m,\; \sum_{m\in
\mathcal{M}_k}b_m\prod_{j\in \mathcal{M}_k\setminus
\{m\}}a_j\right).$$       

Using this, and using the equality $\mathcal{M}_{k+1}=\mathcal{M}_{k}
\cup \{k+1\}$, it is easy to obtain (\ref{Lemma 1.}) for $k+1$ -
element set $\mathcal{M}_{k+1}=\{1,\ 2,\ ... \ ,\ k+1\}$:
$$\bigotimes_{m\in {\mathcal{M}_{k+1}}}(a_m,\ b_m)=%
\bigotimes_{m\in {\mathcal{M}_k}}(a_m,\ b_m)\otimes(a_{m+1},\ b_{m+1})=$$       
$$\left(\prod_{m\in \mathcal{M}_{k+1}}a_m,\ b_{k+1}\prod_{m\in \mathcal{M}_k}a_m+\sum_{m\in \mathcal{M}_k}b_m\prod_{j\in \mathcal{M}_{k+1}\setminus
 \{m\}}a_j \right)$$                                                       
$$=\left(\prod_{m\in \mathcal{M}_{k+1}}a_m,\; \sum_{m\in
\mathcal{M}_{k+1}}b_m\prod_{j\in \mathcal{M}_{k+1}\setminus
\{m\}}a_j\right),$$ %
which proves the lemma.

\end{IEEEproof}

\end{lemma}


\section{The Entropy Message Passing Algorithm}

Let $w(\mathbf{x})$ be a multivariate function whose codomain is
an entropy semiring $\mathcal{K}$ and let the factorization
\begin{eqnarray}
\label{w(x)}
w(\mathbf{x})=\bigotimes_{m\in{\mathcal{M}}}w_m(\mathbf{x_m})
\end{eqnarray}
hold for a set of indices $\mathcal{M}$, where each factor
$w_m(\mathbf{x}_m)$ depends on $\mathbf{x}_m \subset \mathbf{x}$
and subsets $\mathbf{x}_m$ cover $\mathbf{x}$. Further, let the
factors have a form
\begin{eqnarray}
\label{w_m}
w_m(\mathbf{x}_m)=\left(f_m(\mathbf{x}_m),f_m(\mathbf{x}_m)g_m(\mathbf{x}_m)\right),
\end{eqnarray}
where $f_m(\mathbf{x}_m)$ and $g_m(\mathbf{x}_m)$ are real
functions which depend on the same set of variables
$\mathbf{x}_m$. Using (\ref{Lemma 1.}) it is easy to obtain:
\begin{eqnarray}
\label{w(x) expanded} w(\mathbf{x})=
\left(\prod_{m\in{\mathcal{M}}}f_m(\mathbf{x}_m),\prod_{m\in{\mathcal{M}}}f_m(\mathbf{x}_m)\sum_{k\in{\mathcal{M}}}g_k(\mathbf{x}_k)\right).
\end{eqnarray}

Hence, if a function has the form (\ref{w(x) expanded}), then it
can be factorized as in (\ref{w(x)}) with factors as in
(\ref{w_m}).

With a fast computation of
\begin{equation}
\label{Total sum for (Z,H)}%
(Z,H)=\bigoplus_\mathbf{x}w(\mathbf{x}),
\end{equation}
we solve two problems:

1) The computation of the expression:
\begin{equation}
\label{Total sum for Z}
Z=\sum_{\mathbf{x}}\prod_{m\in{\mathcal{M}}}f_m(\mathbf{x}_m),
\end{equation}
which is the normalization problem considered in section II and

2) The computation of the expression:
\begin{equation}
\label{Total sum for H} H=\sum_{\mathbf{x}}
\prod_{m\in{\mathcal{M}}}f_m(\mathbf{x}_m)\sum_{k\in{\mathcal{M}}}g_k(\mathbf{x}_k),
\end{equation}
which is the general form of the different problems described in
the next section. These problems are the key motivation for our
work.

If the factor graph corresponding to $w(\mathbf{x})$ has a tree
structure, the computation (\ref{Total sum for (Z,H)}) can be
performed by message passing over the entropy semiring. Note that
the factor graph of $w(\mathbf x)$ has the same topology as the
function
\begin{equation}
f(\mathbf{x})=\prod_{m\in{\mathcal{M}}}f_m(\mathbf{x}_m);
\end{equation}%
in particular, the factor graph of $w(\mathbf x)$ is cycle free if
and only if the factor graph of $f(\mathbf x)$ is cycle free. To
perform the summation (\ref{Total sum for (Z,H)}), we follow the
procedure from section II - we calculate the marginal:
\begin{equation}
W_n(x_n)=\bigoplus\limits_{\mathbf{x}\setminus x_n}w(\mathbf x)
\end{equation}%
for a variable $x_n$, and subsequently we obtain the total sum by:
\begin{equation}
\bigoplus_{\mathbf{x}}w(\mathbf{x})=\bigoplus_{x_n}W_n(x_n).
\end{equation}

In the following paragraphs, we formalize the discussion by using
the Entropy Message Passing (\emph{EMP}) algorithm:

\subsubsection{Initialization}
Set the messages from all variable and factor nodes in leaves to:
\begin{eqnarray}
\label{EMP Initialization 1} q_{n\rightarrow m}(x_n)=(1,0),
\end{eqnarray}
\begin{eqnarray}
\label{EMP Initialization 2} r_{m\rightarrow
n}(x_n)=\left(f_m(x_n),f_m(x_n)g_m(x_n)\right).
\end{eqnarray}
\subsubsection{Induction}
After receiving the messages from all descendants, compute the
messages to the parents for all variable and factor nodes in the
tree:
\begin{eqnarray}
\label{EMP Induction 1} q_{n\rightarrow m}(x_n)=\bigotimes_ {m'\in
\mathcal{N}(n)\setminus m}r_{m'\rightarrow n}(x_n),
\end{eqnarray}
\begin{eqnarray}
\label{EMP Induction 2} r_{m\rightarrow
n}(x_n)=\bigoplus_{\mathbf{x}_m\setminus
x_n}\left(f_m(\mathbf{x}_m),f_m(\mathbf{x}_m)g_m(\mathbf{x}_m)\right)\nonumber\\
\bigotimes_ {n'\in \mathcal{N}(m)\setminus n}q_{n'\rightarrow
m}(x_{n'}).
\end{eqnarray}
\subsubsection{Termination}
At the root, compute the marginal value and the total sum:
\begin{eqnarray}
\label{EMP Termiantion} \left(Z_n(x_n),\ H_n(x_n)\right)=
\bigotimes_ {m\in \mathcal{N}(n)} r_{m\rightarrow n}(x_n),
\end{eqnarray}
\begin{eqnarray}
\label{Calculation of total sum}
(Z,H)=\bigoplus_{x_n}\left(Z_n(x_n),\ H_n(x_n)\right).
\end{eqnarray}

The \emph{EMP} has the same asymptotic computational complexity as
the \emph{SPA}, since addition and multiplication in an entropy
semiring are realized via addition and multiplication of the real
numbers. The precise complexity estimates of the message passing
algorithms can be found in \cite{Aji-McEliece: GDL}.

Note that in the first component, the entropy semiring works like
ordinary addition and multiplication; in consequence, \emph{EMP}
(i.e., sum-product message passing over the entropy semiring)
contains ordinary sum-product message passing in the first
component.


\section{Applications}

In this section we show how \emph{EMP} applies to the entropy
computation and the optimization techniques such as expectation
maximization and gradient ascent algorithm.


\subsection {Entropy computation of a partially observed probabilistic model}

In this section we show how \emph{EMP} can be used for an
efficient computation of the state sequence entropy of the
partially observed probabilistic models. The algorithm for such
computation has previously been proposed in \cite{Hernando:
Entropy of HMM}, but only for the chain structured models.
Applying the \emph{EMP}, we can generalize this algorithm to the
arbitrary probabilistic model the factor graph of which has no
cycles.

Let a partially observed model be given with the probability
distribution $P(\mathbf x,\mathbf y)$, where $\mathbf x= \{x_1,\
\dots ,\ x_n\}$ denotes a hidden variable sequence of length $n$ and
$\mathbf y= \{y_1,\ \dots \ , y_m\}$ denotes an observation sequence
of length $m$. The entropy of the model $P(\mathbf x,\mathbf y)$ for the given
sequence of the observation is given with:
\begin{eqnarray}
\label{Definition of entropy}
H(X|Y=\mathbf{y})=-\sum_{\mathbf{x}}P(\mathbf{x} |
\mathbf{y})\log_2 P(\mathbf{x} | \mathbf{y}).
\end{eqnarray}
By use of the Bayes theorem and the additivity of a logarithm,
this expression can be transformed to:\vskip 0.05mm
$$\hskip -5cm H(X|Y=\mathbf{y})=$$
\vskip -0.5cm {\begin{eqnarray} \label{Transformation of entropy}
-\frac{\sum_{\mathbf{x}} P(\mathbf{x},\mathbf{y})\log_2
P(\mathbf{x},\mathbf{y})}{\sum_{\mathbf{x}}
P(\mathbf{x},\mathbf{y})}+\log_2 \sum_{\mathbf{x}}
P(\mathbf{x},\mathbf{y}).
\end{eqnarray}}
Note that the probability distribution $P(\mathbf{x},\mathbf{y})$
can be considered as a function depending only on the vector
variable $\mathbf x$, since $\mathbf y$ is observed and can be
treated as a constant.

Let $P(\mathbf{x},\mathbf{y})$ factorize as
\begin{eqnarray}
\label{Factorization for p(x,y)}
P(\mathbf{x},\mathbf{y})=\prod_{m\in{\mathcal{M}}}f_m(\mathbf{x}_m),
\end{eqnarray}
that corresponds to a cycle-free factor graph. If
$g_m(\mathbf{x}_m)=\log_2 f_m(\mathbf{x}_m)$, the expression
(\ref{Transformation of entropy}) can be written as:
\begin{eqnarray}
\label{Final transformation of entropy} H(X |
Y=\mathbf{y})=-\frac{H}{Z}+\log_2 Z,
\end{eqnarray}
where
\begin{eqnarray}
Z=\sum_{\mathbf{x}}\prod_{m\in{\mathcal{M}}}f_m(\mathbf{x}_m)
\end{eqnarray}
and
\begin{eqnarray}
H=\sum_{\mathbf{x}}\prod_{m\in{\mathcal{M}}}f_m(\mathbf{x}_m)\sum_{k\in{\mathcal{M}}}g_k(\mathbf{x}_k).
\end{eqnarray}
The previous expressions have the form (\ref{Total sum for Z}) and
(\ref{Total sum for H}) and can be computed with the \emph{EMP},
which solves the problem of efficient computation of the model
entropy.

\subsection{Iterative optimization techniques}

Suppose we wish to find
\begin{equation}
\label{Theta max} \hat\Theta_{max}=\arg\max_\Theta p(\Theta)
\end{equation}
with a parameter $\Theta$ taking values from $\mathcal{R}$ or
$\mathcal{R}^k$. We assume that $p(\Theta)$ is the marginal of a
real-valued nonnegative function $p(\mathbf{x},\Theta)$:
\begin{eqnarray}
\label{u(theta)} p(\Theta)=\sum_{\mathbf{x}}p(\mathbf{x},\Theta).
\end{eqnarray}

In this section we consider two popular procedures for solving the
problem (\ref{Theta max}) - the Expectation Maximization
(\emph{EM}) and the gradient ascent algorithm. Both algorithms
seek the solution iteratively with the parameter $\Theta$ being
estimated in each iteration. In the following paragraphs we show
how \emph{EMP} can be used for the computations which appear here.
We suppose that if
\begin{eqnarray}
\label{u(x theta)} p(\mathbf{x},\Theta)=\prod\limits_{m \in
\mathcal M} p_{m}(\mathbf{x}_{m},\Theta)
\end{eqnarray}
is considered as the function of $\mathbf x$ only with $\Theta$
fixed, its factor graph is a tree, similarly as in the previous
papers which consider the \emph{EM} algorithm from the message
passing point of a view \cite{Eckford: MUD with GM}-\cite{Dauwels:
EM as MP I}.


\subsubsection{\textbf{The expectation maximization algorithm}}

The \emph{EM} algorithm \cite{Bishop: book}, \cite{Dauwels: EM as
MP I} attempts to compute (\ref{Theta max}) as follows:

\begin{enumerate}
\item
Choose an initial setting for the parameters $\Theta^\text{old}$.

\item
\emph{E}-step: Evaluate  $p(\mathbf{x}, \Theta^\text{old})$.

\item
\emph{M}-step: Evaluate $\Theta^\text{new}$ given by
\begin{equation}
\label{M step}%
\Theta^\text{new}=%
\arg\max_{\Theta} Q (\Theta, \Theta^\text{old})
\end{equation}
where
\begin{equation}
\label{Q(Theta,ThetaOld)}%
Q(\Theta,\Theta^\text{old})=%
\sum_{\mathbf{x}}%
p(\mathbf{x},\Theta^\text{old})\log p(\mathbf{x},\Theta).
\end{equation}

\item While the convergence criterion is not satisfied, let $\Theta^\text{old} = \Theta^\text{new}$ and return to step
2.
\end{enumerate}%
The \emph{M}-step is usually performed by solving the equation
\begin{equation}
\label{grad u(x theta)}%
\nabla_\Theta Q(\Theta,\Theta^\text{old})=0,
\end{equation}%
where $\nabla_\Theta$ denotes the gradient operator. After
substituting (\ref{u(x theta)}) in (\ref{Q(Theta,ThetaOld)}), the
expression (\ref{grad u(x theta)}) can be transformed into

\begin{equation}
\label{grad u(x theta) working}%
\sum_{\mathbf{x}}%
\prod_{m \in \mathcal M} p_{m}(\mathbf{x}_{m},\Theta^\text{old})%
\sum_{k \in \mathcal M} \nabla_\Theta \log
p_k(\mathbf{x}_{k},\Theta) = 0.
\end{equation}

It can be shown that the \emph{EM} algorithm always leads to a
solution. Nevertheless, it becomes computationally demanding as
the number of the steps required for its convergence and
dimensionality of $\mathbf x$ grow. Yet, this problem can
efficiently be solved with the \emph{EMP} when gradients of the
logarithms of the factors in (\ref{u(x theta)}) linearly depend on
$\Theta$, i.e.
\begin{equation}
\label{grad log u(x theta)}%
\nabla_\Theta \log p_k(\mathbf{x}_k, \Theta) = u_k(\mathbf x_k)
\cdot \Theta + v_k(\mathbf x_k) \cdot \Lambda,
\end{equation}
where $\Lambda$ is a constant vector of the same dimensionality as
$\Theta$ (see  \cite{Eckford: FG EM}, \cite{Loeliger: FG for SP}
and \cite{Ronen: DT EM} for the examples). Accordingly, the
solution of (\ref{grad u(x theta) working}) has the form:
\begin{equation}
\label{Theta} \Theta=-\frac{H_a}{H_b} \cdot \Lambda,
\end{equation}
where
\begin{equation}
\label{H_a}
H_a=\sum_{\mathbf{x}}%
\prod_{m \in \mathcal M} p_{m}(\mathbf{x}_{m},\Theta^\text{old})%
\sum_{k \in \mathcal M} u_k(\mathbf x_k)
\end{equation}
and
\begin{equation}
\label{H_b}
H_b=\sum_{\mathbf{x}}%
\prod_{m \in \mathcal M} p_{m}(\mathbf{x}_{m},\Theta^\text{old})%
\sum_{k \in \mathcal M} v_k(\mathbf x_k).
\end{equation}

The expressions for $H_a$ and $H_b$ can efficiently be computed
with the \emph{EMP} algorithm since both can be derived from
(\ref{Total sum for H}) by the settings
$f_{m}(\mathbf{x}_{m})=p_{m}(\mathbf{x}_{m},\Theta^\text{old})$
and $g_{k}(\mathbf{x}_{k})=u_k(\mathbf{x}_k)$ for $H_a$ and
$f_{m}(\mathbf{x}_{m})=p_{m}(\mathbf{x}_{m},\Theta^\text{old})$
and $g_k(\mathbf{x}_k)=v_k(\mathbf{x}_k)$ for $H_b$.


\subsubsection{\textbf{The gradient ascent algorithm}}

The previously described procedure for parameter estimation can be
applied when the linear dependence (\ref{grad log u(x theta)})
holds, but when the dependency is nonlinear, the analytic solution
for the \emph{M}-step does not exist in general. Instead, we can
apply the gradient ascent algorithm \cite{Loeliger: FG for
SP}-\cite{Loeliger: Some remarks on FG} to solve the optimization
problem (\ref{Theta max}). The gradient ascent seeks the maximum
of real nonnegative differentiable function $p(\Theta)$ by an
iterative process:
\begin{equation}
\label{Steepest descent}%
\Theta_{i+1}= \Theta_i%
+\nabla_\Theta p(\Theta)|_{\Theta_i},
\end{equation}
where $\nabla_\Theta p(\Theta)|_{\Theta_i}$ denotes the gradient
of $p(\Theta)$ at the point $\Theta_i$. Since $p(\Theta)$ is given
by the marginal (\ref{u(theta)}), the gradient can be written as
\begin{eqnarray}
\nabla_\Theta p(\Theta)|_{\Theta_i}=
\sum_{\mathbf{x}}\nabla_\Theta p(\mathbf{x},\Theta) |_{\Theta_i}.
\end{eqnarray}
If we apply Leibniz's rule to the factorization (\ref{u(x
theta)}), the previous expression becomes:
\begin{equation}
\label{Expression for teta}
\nabla_\Theta p(\Theta)|_{\Theta_i}=%
\sum_{\mathbf{x}} \prod_{m \in \mathcal{M}}f_{m}(\mathbf{x}_{m})
\sum_{k \in \mathcal M} g_{k}(\mathbf{x}_{k}),
\end{equation}
where
\begin{eqnarray}
\label{Factors for f in SD}%
f_{m}(\mathbf{x}_{m})= p_{m}(\mathbf{x}_{m},\Theta_i)
\end{eqnarray}
and
\begin{eqnarray}
\label{Factors for g in SD}
g_{k}(\mathbf{x}_{k})=\frac{\nabla_\Theta
p_{k}(\mathbf{x}_{k},\Theta)}{p_{k}(\mathbf{x}_{k},\Theta)}
\mid_{\Theta=\Theta_i}.
\end{eqnarray}
Again, we have the expression of the form (\ref{Total sum for H}),
so the gradient (\ref{Expression for teta}) can be evaluated with
the \emph{EMP} algorithm.

The gradient ascent can also be used for the \emph{M}-step of the
\emph{EM} algorithm as in \cite{Dauwels: SD as MP}, \cite{Seq Algs
based on KL measure} and \cite{On line estimation based on KLD}.
In this case, (\ref{Q(Theta,ThetaOld)}) should be maximized by the
iterative procedure:
\begin{equation}
\label{SD for Q}%
\Theta_{i+1}= \Theta_i%
+\nabla_\Theta Q(\Theta,\Theta^\text{old})|_{\Theta_i}.
\end{equation}
The computation that appears here can also be performed with the
\emph{EMP}, since the gradient of $Q(\Theta,\Theta^\text{old})$
reduces to (\ref{Expression for teta}) for
\begin{eqnarray}
\label{Factors for f in SD}%
f_{m}(\mathbf{x}_{m})= p_{m}(\mathbf{x}_{m},\Theta^\text{old})
\end{eqnarray}
and
\begin{eqnarray}
\label{Factors for g in SD}
g_{k}(\mathbf{x}_{k})=\frac{\nabla_\Theta
p_{k}(\mathbf{x}_{k},\Theta)}{p_{k}(\mathbf{x}_{k},\Theta)}
\mid_{\Theta=\Theta_i},
\end{eqnarray}
which can easily be shown.

\section{Conclusion}

Building on previous work \cite{Cortes: KLD betwen automata},
\cite{Hernando: Entropy of HMM}, we have proposed a new general
message passing algorithm for factor graphs, which we call entropy
message passing (\emph{EMP}). \emph{EMP} may be viewed as a new
version of sum-product message passing over the entropy semiring.
The following applications of \emph{EMP} have been demonstrated:
1) the computation of the entropy of an observed hidden Markov
model with fixed observations (in this application, \emph{EMP} is
essentially the algorithm of \cite{Cortes: KLD betwen automata}),
2) expectation maximization and 3) gradient-ascent algorithms.

As a version of sum-product message passing, \emph{EMP} gives
exact results only for factor graphs without cycles. Nevertheless,
the algorithm can be applied (without guarantees) also to factor
graphs with cycles, where it might give good empirical results in
some cases.

\section{ACKNOWLEDGMENT}

The authors would like to thank thank the anonymous reviewers for
their valuable comments and suggestions that improved the
presentation of this paper.

\end{document}